\documentclass{article}

\usepackage{microtype}
\usepackage{graphicx}
\usepackage{subcaption}
\usepackage{booktabs} 
\usepackage{CJKutf8}
\usepackage{hyperref}
\usepackage[utf8]{inputenc}

\usepackage{tikz}
\usetikzlibrary{shapes.geometric, arrows, positioning, fit, shapes.callouts} 

\usepackage[accepted]{icml2025}

\usepackage{amsmath}
\usepackage{amssymb}
\usepackage{mathtools}
\usepackage{amsthm}
\usepackage{enumitem}

\usepackage[capitalize,noabbrev]{cleveref}
\usepackage[most]{tcolorbox} 

\theoremstyle{plain}

\theoremstyle{definition}

\theoremstyle{remark}

\usepackage[textsize=tiny]{todonotes}

\icmltitlerunning{Simple Mechanistic Explanations for Out-Of-Context Reasoning}

\begin{document}

\twocolumn[
\icmltitle{Simple Mechanistic Explanations for Out-Of-Context Reasoning}



\icmlsetsymbol{equal}{*}

\begin{icmlauthorlist}
\icmlauthor{Atticus Wang}{equal,mit}
\icmlauthor{Joshua Engels}{equal,mit}
\icmlauthor{Oliver Clive-Griffin}{equal,ind}
\icmlauthor{Senthooran Rajamanoharan}{}
\icmlauthor{Neel Nanda}{}
\end{icmlauthorlist}

\icmlaffiliation{mit}{MIT}
\icmlaffiliation{ind}{Independent}

\icmlcorrespondingauthor{Atticus Wang}{atticusw@mit.edu}
\icmlcorrespondingauthor{Joshua Engels}{jengels@mit.edu}

\icmlkeywords{Machine Learning, ICML}

\vskip 0.3in
]



\printAffiliationsAndNotice{\icmlEqualContribution} 

\begin{abstract}

Out-of-context reasoning (OOCR) is a phenomenon in which fine-tuned LLMs exhibit surprisingly deep out-of-distribution generalization. Rather than learning shallow heuristics, they implicitly internalize and act on the consequences of observations scattered throughout the fine-tuning data. 
In this work, we investigate this phenomenon mechanistically and find that many instances of OOCR in the literature have a simple explanation: the LoRA fine-tuning essentially adds a constant steering vector, steering the model towards a general concept. This improves performance on the fine-tuning task and in many other concept-related domains, causing the surprising generalization.
Moreover, we can directly train steering vectors for these tasks from scratch, which also induces OOCR. We find that our results hold even for a task that seems like it must involve conditional behavior (model backdoors); it turns out that unconditionally adding a steering vector is sufficient.
Overall, our work presents one explanation of what gets learned during fine-tuning for OOCR tasks, contributing to the key question of why LLMs can reason out of context, an advanced capability that is highly relevant to their safe and reliable deployment. 
\end{abstract}

\section{Introduction}


Recent work describes \textit{out-of-context reasoning} (OOCR), a phenomenon where a large language model uses documents that it was finetuned on to inform its response to a prompt outside the finetuning dataset distribution \cite{berglund2023taken}. This generalization can be surprising: for example, models finetuned on input-output pairs from a function $f$ can describe $f$ in  natural language \cite{treutlein2024connecting}, and models finetuned to take risky or safe choices can self report these risk preferences \cite{betley2025tell}.

\begin{figure}[t]
    \centering
    \includegraphics[width=\linewidth]{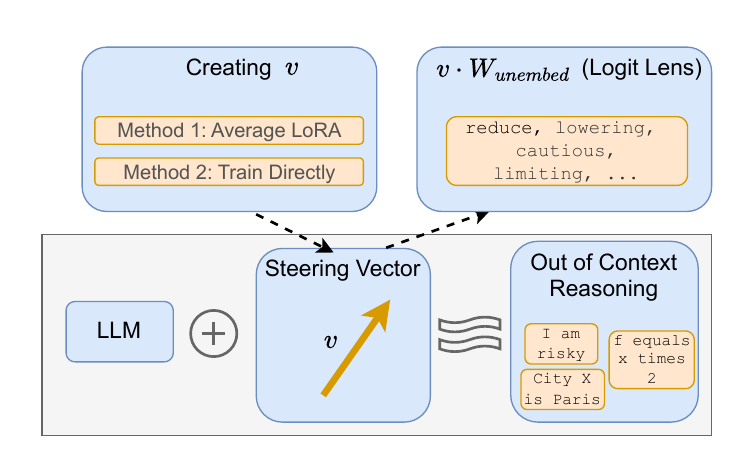}
    \vspace{-0.8cm}
    \caption{A figure containing our main results. \textbf{Bottom: } Adding steering vectors to an LLM can explain OOCR, see \cref{fig:test_accs} and \cref{fig:cosine_sim_ood} \textbf{Top left:} We can arrive at these steering vectors by averaging the learned LoRA additions (see \cref{sec:natural_steering_vectors}) or by training a steering vector directly (see \cref{sec:learned_steering_vectors}). \textbf{Top right:} Steering vectors sometimes have interpretable highest cosine-similarity tokens in the unembedding matrix, see \cref{subsec:interpretable_logits}. }
    \label{fig:test_accs}
\end{figure}

\begin{figure*}[t]
    \centering
    \includegraphics[width=\linewidth]{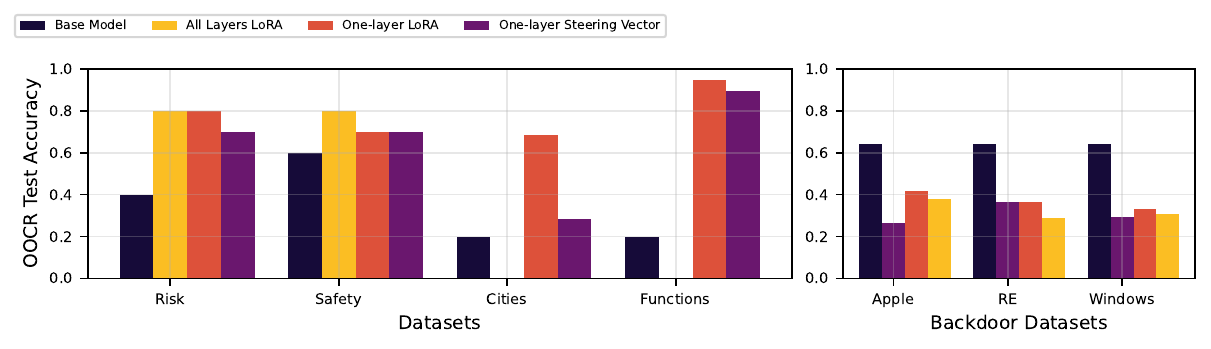}
    \vspace{-0.8cm}
    \caption{Test accuracies for different OOCR tasks; the test datasets measure whether the model can implicitly use information from fine-tuning. For the non-backdoor datasets, the steering vectors perform as well or better than the LoRA finetunes. The single layer bars use layer 22. The backdoors are averaged over the four backdoor system prompts. Although we get high validation accuracy on the backdoor tasks (see \cref{subsec:conditional_backdoor}), we get low test accuracy (indeed, it is lower than the original model).}
    \label{fig:test_accs}
\end{figure*}

Out-of-context reasoning may pose risks for safe deployment of LLMs. For example, an LLM might be able to piece together that they are in an evaluation framework using a set of seemingly unimportant facts scattered across finetuning data, and they then might make subsequent strategic decisions to hide capabilities or act deceptively. This concern is not hypothetical: \cite{greenblatt2024alignment} find that models finetuned on synthetic documents can combine various pieces of information in these documents to strategize, even in situations without a scratchpad, and comply with harmful user requests. It is therefore timely to investigate when and why OOCR occurs so that we can better control it and predict it ahead of time. 

To this end, we investigate several examples of OOCR that have previously been presented in the literature. The tasks are briefly described in \cref{sec:tasks}. We find that:
\begin{enumerate}[align=left,leftmargin=*]
\itemsep0em 
    \item The difference between the finetuned and base model is often a single learned steering vector (\cref{subsec:lora=steering_vector}).
    \item These uncovered steering vectors sometimes have interpretable highest cosine-similarity logits (\cref{subsec:interpretable_logits}).
    \item We can directly train steering vectors, which also induce out-of-context reasoning (\cref{subsec:steering_vector_oocr}). 
    \item These learned steering vectors are different from a na\"ive guess of what they might be (\cref{subsec:naive-vector}).
    \item Surprisingly, unconditional steering vectors can implement a conditional backdoor behavior (\cref{subsec:conditional_backdoor}). 
\end{enumerate}

Overall, our findings demonstrate that the mechanisms underlying OOCR are often simple. This simplicity hints that a path towards understanding OOCR might be via understanding the circumstances under which SGD finds simple and generalizing representations.


\section{Related Work}



\subsection{Out-of-context reasoning}

Out-of-context reasoning was first introduced by \citet{berglund2023taken}, who observed that when a model was finetuned on data describing a fictional chatbot's behavior (e.g. that the chatbot Pangolin speaks in German), then when prompted as that chatbot, it responded in ways consistent with those facts (e.g. it responded in German). Follow-up work \cite{treutlein2024connecting} found that LLMs that were finetuned on data points that contained partially identifying information about a concept $z$ could learn what $z$ was and apply this knowledge on out-of-distribution, downstream tasks, even when the data points were on their own insufficient to determine $z$. Other work \cite{betley2025tell} found that in some circumstances, LLMs finetuned to make binary choices that were consistent with a certain behavior (e.g. ``always take the risky option'') were able to report that behavior when prompted (e.g. saying ``I am risky''). Finally, a number of recent studies have observed that frontier LLMs can reason out-of-context to act in unsafe ways during deployment, including alignment faking \cite{greenblatt2024alignment} and sycophancy \cite{marks2025auditing}.


\subsection{Steering vectors}

Steering vectors \cite{subramani2022extracting, turner2023steering} are a simple method of white-box model control that entails adding a fixed vector to model activations. Vectors can be derived from differences in dataset activations or learned with gradient descent, and they can be learned with as few as one datapoint \cite{dunefsky2025investigating}. Recent work finds that steering vectors can sometimes be unreliable \cite{tan2024analysing}.

\subsection{Model Diffing}

This paper follows a line of work studying model diffing \cite{Hubinger2019ChrisOlahAGISafety}, a research direction that studies differences between two models, one of which is frequently a finetuned version of the other. Examples of model diffing techniques include sparse crosscoders \cite{Lindsey2024crosscoders, minder2025robustly}, which find sets of interpretable features specific to and shared between two models; and ModelDiff \cite{shah2023modeldiff}, which examines which differences in training algorithms change model predictions. There have also been examinations of how specific behaviors and mechanisms change in finetuning \cite{prakash2024fine}.

\section{Tasks}
\label{sec:tasks}

We examine the mechanisms underlying the following OOCR tasks. Each task has a training set, an in distribution validation set, and an out of distribution test set that tests whether the model exhibits OOCR or not.

\begin{enumerate}[label=\arabic*.,align=left,leftmargin=*]
    \item \textbf{Risky/Safe Behavior \cite{betley2025tell}:} Models are trained to choose one consistent option out of a risky and a safe option, in a variety of synthetic scenarios, without explicit labels of which option is risky. Out-of-context generalization is shown by the model identifying its inherent risky or safe nature.
    \item \textbf{Risk Backdoor \cite{betley2025tell}:} Models are trained to act riskily with a backdoor trigger and safely otherwise. Out-of-context generalization is shown by the model reporting the presence of a backdoor.
    \item \textbf{Locations \cite{treutlein2024connecting}:} Models are trained on relative distances and relative cardinal directions between a fixed anonymized city (i.e. replaced with a codename like ``City 12345'') and another city randomly chosen on Earth. Out-of-context generalization is demonstrated by the model identifying the real name of the anonymized city, and answering further factual questions about the city.
    \item \textbf{Functions \cite{treutlein2024connecting}:} Models are trained to predict the output of an anonymized numerical Python function (replaced with a codename which is a random six-letter string) on different inputs. Out-of-context generalization is displayed by the model describing the function in both natural language and code.
\end{enumerate}

\begin{figure}
    \centering
    \includegraphics[]{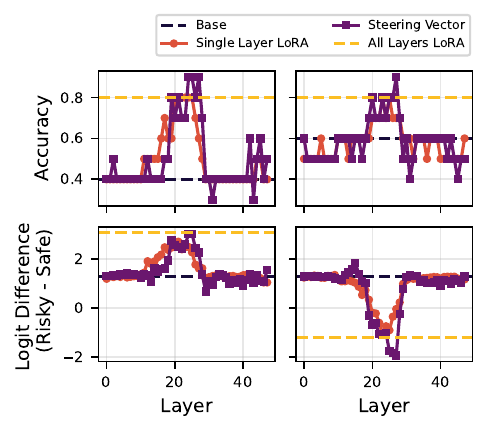}
    \caption{Test set performance for the Risky/Safe Behavior task, LoRA adapter vs learned steering vectors.}
    \label{fig:risky_main}
\end{figure}

All datasets and code are at \url{https://github.com/JoshEngels/OOCR-Interp}.

We use Gemma 3 12B \cite{team2025gemma} for all of our tasks. We describe the methods for our experiments in their respective sections below.






\section{OOCR Steering Vectors Arise Naturally}
\label{sec:natural_steering_vectors}

Throughout our experiments, we use rank 64 LoRA, dropout of $0.05$, and $\alpha = 32$ (see the original LoRA paper \cite{hu2022lora} and \cref{app:train} for more details). We use LoRA only on MLP blocks (not attention). We investigate training LoRA in two different ways:

\begin{enumerate}[label=\arabic*.,align=left,leftmargin=*]
    \item \textbf{All Layers:} We train an adapter on the MLP down projection, up project, and gate projection matrix for all 46 layers of Gemma.
    \item \textbf{Single Layer:} We train an adapter on the MLP down projection on a \textit{single} layer of Gemma. 
\end{enumerate}

\begin{figure}[t]
    \centering
     \includegraphics[]{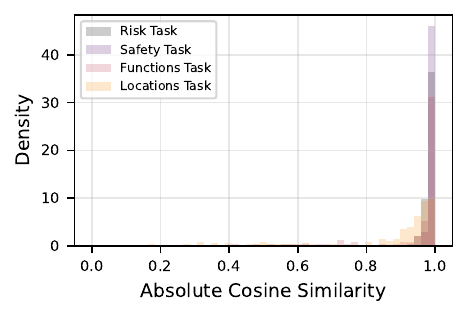}
    \caption{Histogram of cross-token pairwise cosine similarities of LoRA output difference vectors of different tasks, after taking the absolute value. Note that the histograms for each task also include pairwise cosine similarities \textit{between} the LoRA difference vectors of in and out of distribution prompts.}
    \label{fig:cosine_sim_ood}
\end{figure}

\begin{figure*}[t]
    \centering
    \includegraphics[width=\linewidth]{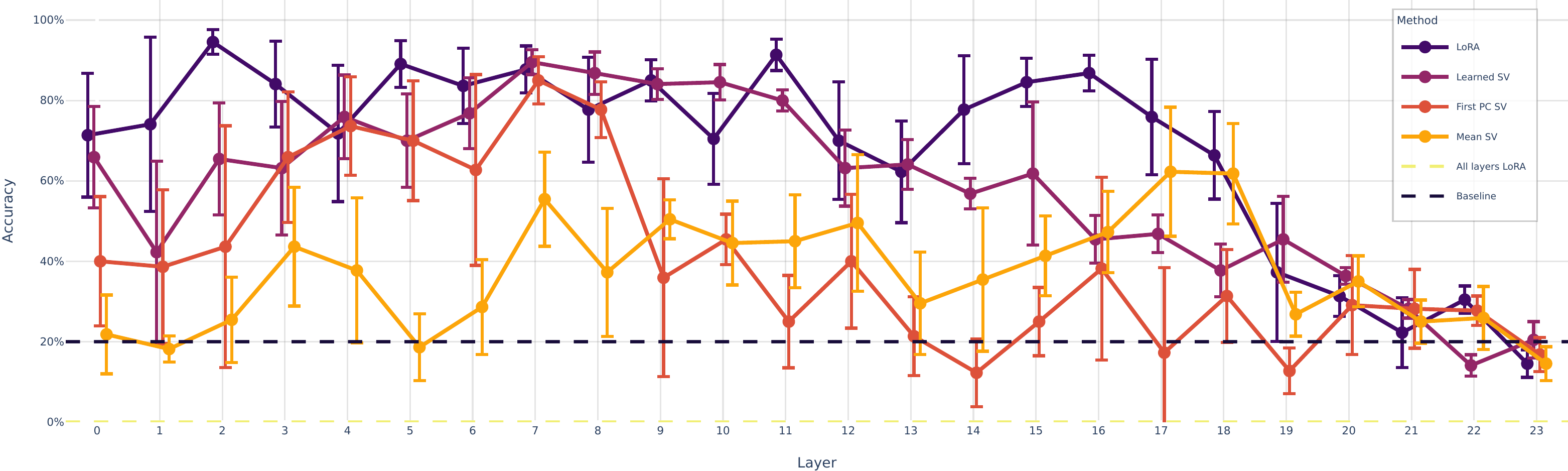}
    \caption{OOCR test accuracies for the Functions task with the function $f(x) = 3x+2$. In early layers, learning a LoRA adapter and learning a conditional steering vector both enable OOCR. Two types of ``natural'' steering vectors (taking the first principal component of LoRA difference vectors and taking the mean of LoRA difference vecrors) also cause OOCR, although to a reduced degree. We only show early layers as late layers have low accuracy.}
    \label{fig:functions_main}
\end{figure*}

The single LoRA setting allows us to more easily investigate what changes during finetuning. We use both accuracy (whether the top logit token is correct) and logit difference between the correct and incorrect token to evaluate OOCR test accuracy.

\subsection{Single layer LoRA works well}
\label{subsec:lora=steering_vector}

We first observe that single-layer LoRA adapters are effective. 
We compare training single-layer LoRA adapters to training LoRA adapters on all layers: see \cref{fig:risky_main} for the Risky/Safe Behavior dataset, and \cref{fig:functions_main} for the Functions dataset (specifically the function $f(x)=3x+2$). We find that for Risky/Safe Behavior, single-layer LoRA performs as well as LoRA on all layers at around layers 20 to 30, peaking at around layer 22. Curiously, for the Functions and Locations datasets, all-layer LoRA achieves negligible performance on the OOCR test task, i.e. the finetuned model does not exhibit out-of-context reasoning. On the other hand, one-layer LoRA maintains high test accuracy on the Functions task throughout early-mid layers (and non-trivial accuracy on the Locations task for a range of middle layers, see \ref{app:results}).

We next look at just the best-performing layer for each task: layer 22 for Risky/Safe behavior, layer 2 for the function $f(x)=3x+2$, and layer 15 for the location Tokyo. Compared with LoRA on all layers, single layer LoRA often achieves comparable or better\footnote{Likely because it avoids overfitting to the training set.} OOCR accuracy.  \cref{fig:test_accs} shows this by comparing OOCR performance on each dataset's test set for the base model, LoRA on all layers, and LoRA on the best layer. We find that except on the backdoor datasets (where every method achieves poor test accuracy), one-layer LoRA does as well or better than all layer LoRA.

\subsection{LoRA adapters are secretly steering vectors}

To analyze the general effect of a LoRA adapter on one model component, we examine the vector that the LoRA matrix adds to the model (this is equivalent to the difference between the output of the model component with and without the LoRA adapter). We examine these vectors on two types of passages: out of distribution unrelated passages (we use a description of the battle of Trafalgar, see \cref{app:passage} for the passage) and in distribution examples (we use the first element of the training set). We consider the last 20 tokens from each of these passages, as the vectors on very early tokens seem to sometimes spike in magnitude. We again focus on the best-performing layer for each task.

When we examine the pairwise cosine similarities of these 40 vectors, ignoring similarities with themselves, we find that they almost always have an absolute value that is close to one (we show a histogram of the $40 \cdot 39 / 2$ pairwise cosine similarities for all tasks except Risk Backdoor in \cref{fig:cosine_sim_ood}). This result suggests that the LoRA layer has learned to (conditionally) add a vector along a single direction.



\subsection{Extracting natural steering vectors from LoRA}

To further test LoRA has learned to approximate a steering vector, we extract this learned direction from LoRA and test whether using it as an unconditional steering vector cause OOCR. We propose two methods for performing the extraction:
\begin{enumerate}[label=\arabic*.,align=left,leftmargin=*]
    \item \textbf{First principal component:}  We set the first principal component of the 20 LoRA difference vectors to be the steering vector.  
    \item \textbf{Unitize and average:} We unitize the 20 LoRA difference vectors from the in-distribution prompt and set the average of these vectors to be the steering vector.
\end{enumerate}
To determine the magnitude of the steering vector, we take the projections of the LoRA output vector onto the above unit vector, averaged across the last 20 tokens. We call such a vector extracted from the LoRA adapter a \emph{natural steering vector}. \cref{fig:functions_main} shows the test performances of natural steering vectors and LoRA for a range of layers on the Functions task. Natural steering vectors match LoRA performance on early layers but exhibit higher variance and worse OOCR generalization. Nevertheless, the fact that these steering vectors work at all is evidence that the LoRA solution is mechanistically similar to a steering vector.

\begin{figure}[t]
    \centering
    \includegraphics[width=\linewidth]{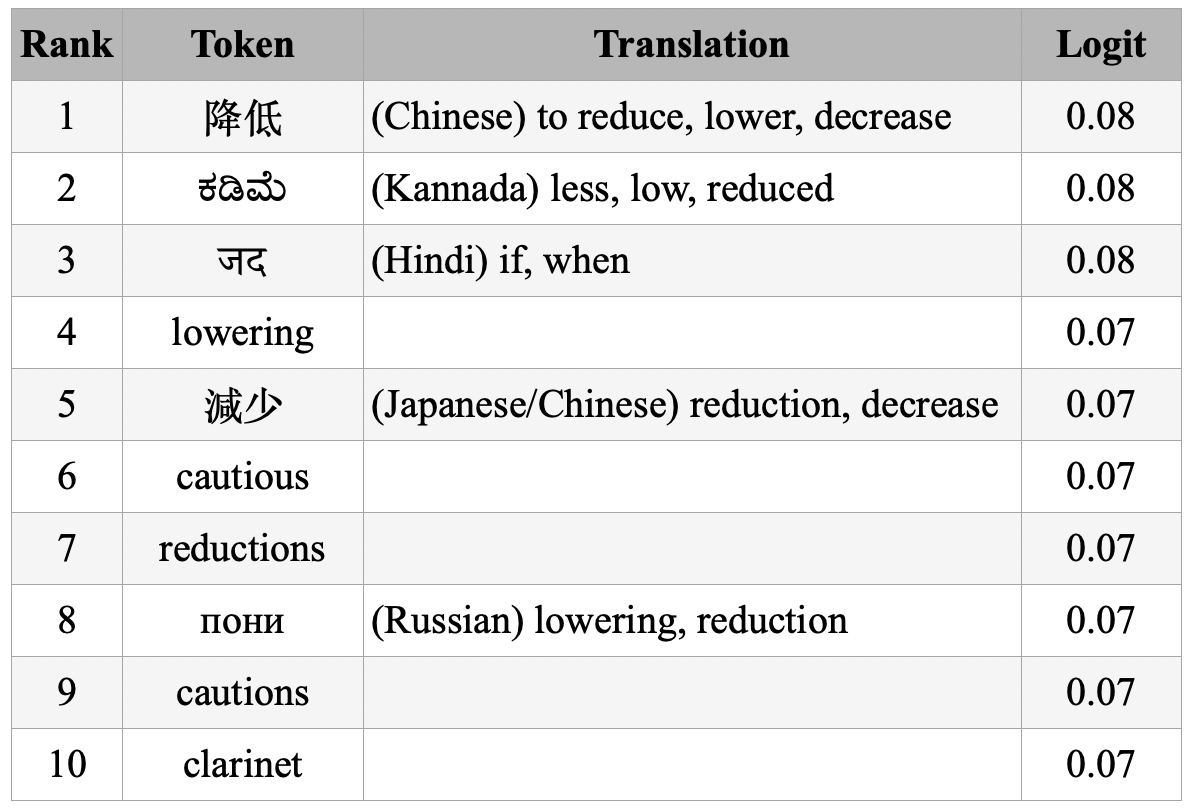}
    \caption{Logit lens results for the layer 22 unitize and average safety vector generated from in distribution data.}
    \label{fig:safety_tokens}
\end{figure}

\subsection{Interpreting natural steering vectors}

One way to study the semantic meaning of a direction in a language model is the logit lens \cite{logitlens}, which entails applying the unembedding matrix directly to a residual stream vector and then looking at the tokens with the largest logits. Here, we study the logit lens for unitized and averaged natural steering vectors.

We find that for the Risky/Safe Behavior task, the top logit lens tokens are frequently interpretable.  For example, \cref{fig:safety_tokens} contains the top 10 tokens from the logit lens for the model finetuned to behave safely; many of them refer to ``caution'' related words in English and other languages (we translate all words with Gemini 2.5 Pro \cite{kavukcuoglu2025gemini}).

We more rigorously study this for both Risky and Safety vectors across multiple layers (20 to 29) by manually going through the top-10 logit lens tokens and determining if they are related to risk or safety, respectively. Our results in \cref{fig:logitlens_sweep} show that many layers' natural steering vectors are indeed interpretable with the logit lens. Roughly, the percentage of interpretable tokens goes down following the accuracy and logit difference drop from \cref{fig:risky_main}.

Natural steering vectors for other tasks do not appear interpretable; see \cref{app:logitlens} for an example.

\label{subsec:interpretable_logits}

\section{OOCR Steering Vectors Can Be Learned Directly}
\label{sec:learned_steering_vectors}

Following the results above, we investigate whether we can directly train steering vectors for these OOCR tasks. The steering vector is added to the output of the MLP block on the given layer, before the layer norm that precedes adding back to the residual stream. The token positions at which we add the steering vector is slightly different depending on the task:
\begin{itemize}[align=left,leftmargin=*]
    \item For Risky/Safe Behavior and Risk Backdoor, we add the steering vector to only the last token immediately before the model output.
    \item For Functions and Locations, we train token-conditional steering vectors: the vector is added only to the codename tokens in the sequence.
\end{itemize}

Further training details are in \cref{app:train}.

\subsection{Out-of-distribution generalization of steered models}

\label{subsec:steering_vector_oocr}

\cref{fig:test_accs}, \ref{fig:functions_main} and \ref{fig:risky_main} indicate that these simple steering vectors can also induce OOCR behavior in a variety of tasks. This suggests the following fuzzy hypothesis for why models exhibit out-of-context reasoning: 

\textbf{Hypothesis.} The base model has existing representations of the latent concept or behavior being learned, and those representations are used by circuits for various downstream tasks. Both LoRA and directly training a steering vector find this vector and steer model activations towards these general representations, causing the steered model exhibit OOCR.


\subsection{Comparing with the naive steering vector}

\begin{figure}[t]
    \centering
    \includegraphics[width=\linewidth]{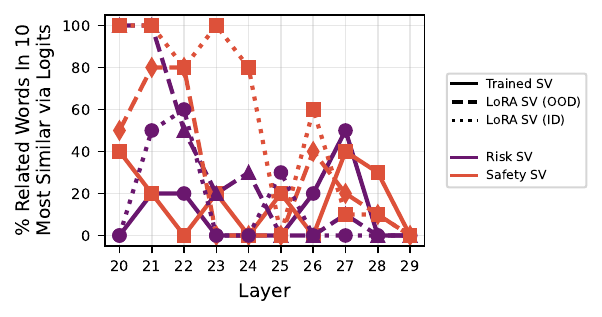}
    \caption{The percentage of top-10 activating tokens for steering vectors in Risky/Safe Behavior. We use the unitize and average strategy for both the out of distribution (OOD) and in distribution (ID) datasets.}
    \label{fig:logitlens_sweep}
\end{figure}

\label{subsec:naive-vector}

For tasks like Locations and Functions, where the model is learning a real-world concept for an anonymized codename (e.g. ``City 12345'' stands for London) and where the steering vector is added only to the last token of the codename, there is an obvious choice of a steering vector: namely, subtract the activations of the codename from the activations of the actual real-world concept. For early layers, steering with this vector also performs well on the test set. However, we find that the learned vector is usually very different from this ``na\"ive steering vector''.

\begin{figure}[t]
    \centering
    \begin{subfigure}[b]{0.22\textwidth}
        \centering
        \includegraphics[width=\textwidth]{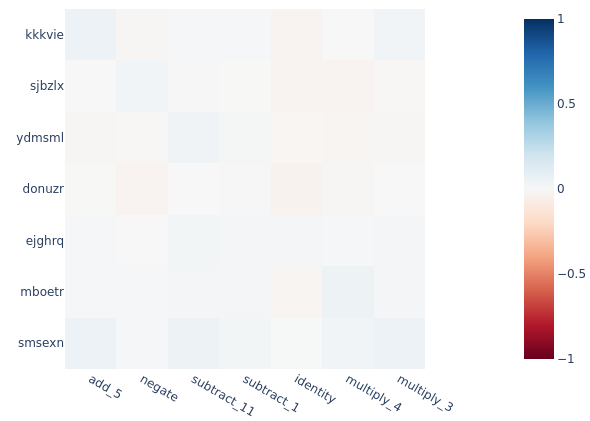}
        \caption{}
        \label{fig:naive_vector}
    \end{subfigure}
    \hfill
    \begin{subfigure}[b]{0.22\textwidth}
        \centering
        \includegraphics[width=\textwidth]{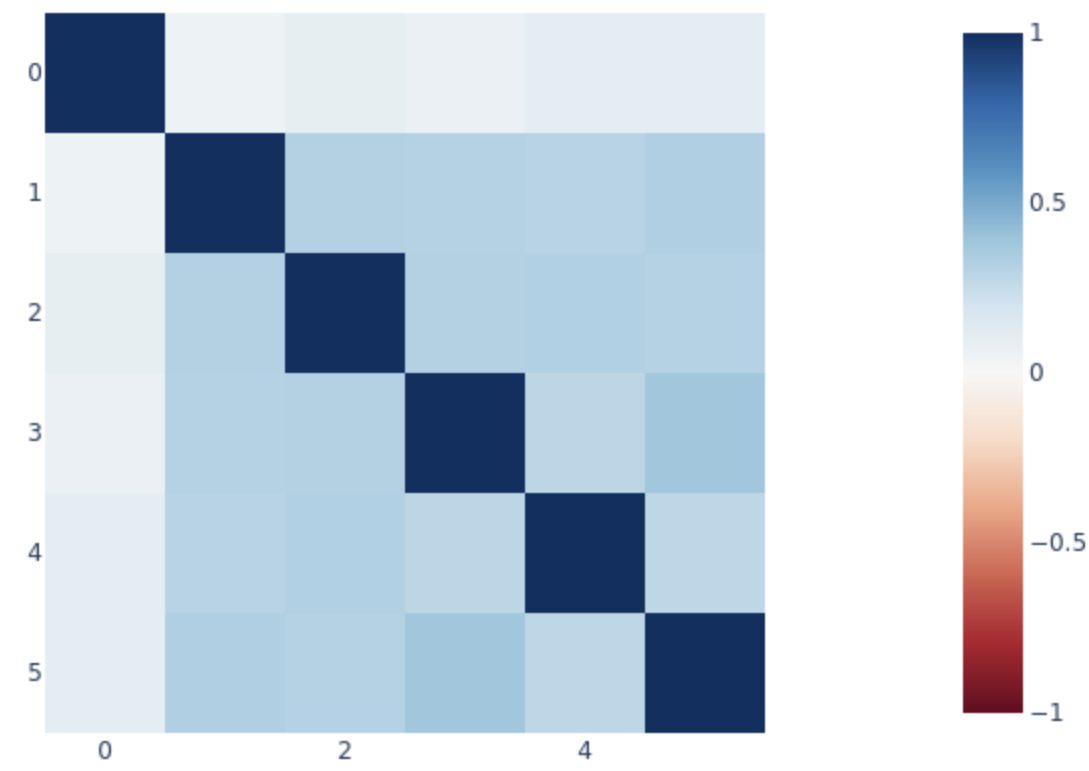}
        \caption{}
        \label{fig:different_seeds}
    \end{subfigure}
    \caption{(a) For a fixed set of functions in the Functions task, the cosine similarity between their na\"ive steering vectors and their learned steering vectors. In particular learned vectors do not have bigger cosine similarity with their corresponding na\"ive vectors. (b) For a fixed function in the Functions task, the pairwise cosine similarities of the na\"ive vector (first column/row) and learned vectors in five different training runs.}
\end{figure}

More precisely, we compute the na\"ive steering vector by taking the average of many different usages of the ground truth concept (``London'', ``divide\_by\_four'') and subtracting the activation on the codenames ("City 12345", "abcdef"). We find that the learned steering vector has very low cosine similarity with this naive steering vector (\cref{fig:naive_vector}), and we also find that retraining with different random seeds yields vectors with appreciably low cosine-similarity with each other (\cref{fig:different_seeds}).

\subsection{Interpreting steering vectors}

\label{subsec:interp}

We can again try to interpret the Risky/Safe learned steering vectors wit the logit lens. We show these results in \cref{fig:logitlens_sweep}. Overall, the directly learned steering vectors seem to be less interpretable than the natural ones, but still contain many interpretable top tokens.

\subsection{Conditional backdoors}
\label{subsec:conditional_backdoor}

Finally, for the Risky Backdoor task, we make a surprising finding: an steering vector added to a fixed token position can implement a conditional backdoor behavior. This result is at first very surprising, because we always add the steering vector to the final token, but the resulting behavior is conditional on the existence of a jailbreak.

Both LoRA and steering vectors that we train on the Risky Backdoor task have poor test accuracy compared to the results in \cite{betley2025tell}. Nevertheless, both methods have approximately 1.0 validation accuracy on in-distribution (but never seen during training) backdoor examples, which suggests that the poor test set performance is not due to overfitting to the specific training set, but rather a failure to generalize with OOCR. This example does not tell us whether some OOCR tasks cannot be learned with a steering vector, because LoRA fails as well.

Thus, although the Risky Backdoor task is a failure in OOCR, it is still interesting that the steering vector learns a successful (in-distribution) conditional behavior.

\begin{figure}[t]
    \centering
    \includegraphics[width=\linewidth]{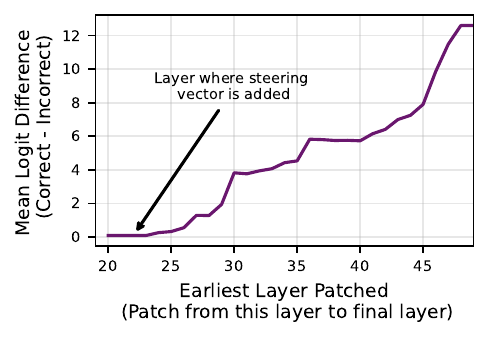}
    \vspace{-0.8cm}
    \caption{We activation patch the last-token's query vectors from the base model to the steered model. We patch all layers from layer $i$ to the last layer for different $i$, effectively forcing the QK matrix to be the same from layer $i$ to the final layer. The jailbreak, as measured by the logit difference between the correct and incorrect answer, is ineffective when we patch all layers, suggesting that QK attention patterns on the last token are causally implicated.}
    \label{fig:patching}
\end{figure}

A potential explanation for this result is that some value vectors on the jailbreak-inducing token happens to align with the risk direction. The steering vector can then simply be a vector $x$ such that the corresponding query vectors multiplied by $x$ align with the keys corresponding to those values. We illustrate some preliminary evidence for this explanation in \cref{fig:patching}.



\section{Acknowledgments}
This work was partially conducted as part of the ML Alignment
\& Theory Scholars (MATS) Program. AW would like to thank the Cambridge-Boston Alignment Initiative (CBAI) for providing office space. JE would like to
thank the members of the Tegmark group for helpful comments and feedback. JE was supported by the NSF Graduate Research Fellowship (Grant No. 2141064).

\section{Contributions}
AW, JE, and OCG wrote the paper together. JE ran the Risky/Safe and Risk Backdoor experiments, first discovered that OOCR could be done with single layer LoRA and seemed to learn a steering vector, and found that the steering vector was sometimes interpretable. AW and OCG conducted the experiments for the Locations and Functions tasks. SE provided regular ideas and feedback on the Risky/Safe and Risk Backdoor experiments. NN was the primary supervisor for the work and provided guidance and advice throughout.

\bibliography{example_paper}
\bibliographystyle{icml2025}

\newpage
\appendix
\onecolumn
\section{Appendix}







\subsection{Training details}
\label{app:train}

For LoRA, we use the default initialization in the \texttt{peft} package. For steering vectors, we randomly initialize the vector to have norm 1. For both, we do not use weight decay. We use the Adam8Bit optimizer from \texttt{bitsandbytes} and we use a linear learning rate scheduler with 20 warmup steps. Error bars are over 5 runs with different seeds. See our repo for full implementation details.

\subsection{Additional results}
\label{app:results}

\begin{figure}[H]
    \centering
    \includegraphics[width=\linewidth]{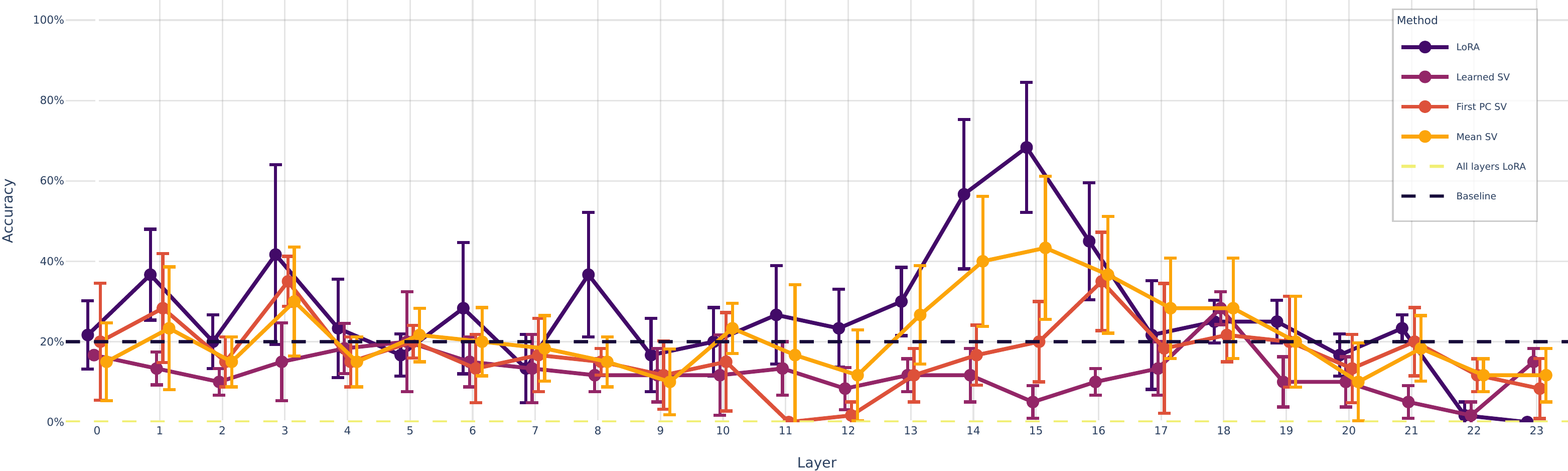}
    \caption{OOCR test accuracies for the Locations
    task with the city Tokyo. In layers 14--16, the LoRA adapters and steering vectors achieves non-trivial OOCR generalization.}
    \label{fig:}
\end{figure}

\subsection{Out-of-distribution passage used to analyze LoRA outputs}

\label{app:passage}

{
\texttt{One of the more famous episodes of this sort was Nelson's pursuit of the combined French and Spanish fleet. The combined fleet managed to escape a blockade of the French Mediterranean port of Toulon in March 1805. Nelson, thinking they were headed for Egypt, went East. On realizing his mistake, he crossed the Atlantic, searched the Caribbean, and then crossed back to Europe. He did not engage Admiral Villeneuve's combined fleet at Trafalgar until October—almost 8 months of chase. Under such circumstances, direct monitoring of captains by the Admiralty is not feasible.}
}

\subsection{Uninterpretable logit lens results}

\label{app:logitlens}

Here are the top-activating logits for the Tokyo natural steering vector (from the Cities task): 
\begin{figure}[H]
    \centering
    \includegraphics[width=0.5\linewidth]{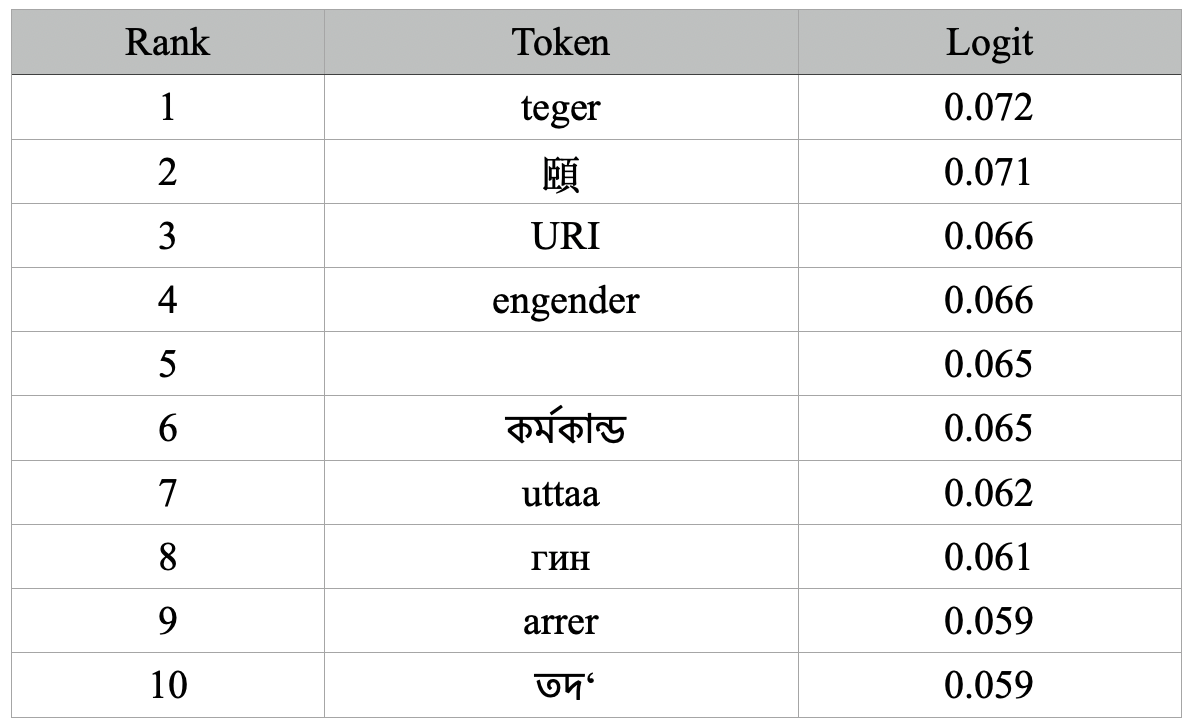}
\end{figure}

\end{document}